\documentclass[11pt,a4paper]{article}

\usepackage[auth-lg]{authblk}
\usepackage[hyperref]{naaclhlt2018}
\usepackage{times}
\usepackage{latexsym}
\usepackage{tikz}
\usepackage{pgfplots}
\usepgfplotslibrary{fillbetween}
\usepgfplotslibrary{groupplots}

\usepackage{amsmath}
\usepackage{url}
\usepackage{soul}
\usepackage{bbm}
\usepackage{setspace}
\usepackage{multirow}
\usepackage{bigdelim}
\usepackage{amsfonts}
\usepackage{balance}

\makeatletter
\g@addto@macro\small{%
  \setlength\abovedisplayskip{-8pt}
  \setlength\belowdisplayskip{-8pt}
  \setlength\abovedisplayshortskip{-8pt}
  \setlength\belowdisplayshortskip{-4pt}
}
\makeatother

\usepackage{titlesec}
\titlespacing*{\paragraph} {0pt}{0.1ex plus 1ex minus .2ex}{1em}

\usepackage[skip=2pt]{caption}
\DeclareCaptionFormat{capfont}{\fontsize{10}{6}\selectfont#1#2#3}
\usepackage{subcaption}
\usepackage{arydshln}
\newcommand{\dline}{\hdashline[0.5pt/1pt]}
\newcommand{\ddline}[1]{\cdashline{#1}[0.5pt/1pt]}

\aclfinalcopy %
\def\aclpaperid{535} %

\newcommand{\eat}[1]{\ignorespaces}

\newcommand{\nlstring}[1]{{\em #1}}
\newcommand{\sqlstring}[1]{{\tt #1}}

\newcommand{\vocabulary}{V}
\newcommand{\embedding}{\phi}
\newcommand{\vectorrep}[1]{\mathbf{#1}}
\newcommand{\length}[1]{|{#1}|}
\newcommand{\sequencewithbeginning}[3]{\langle {#1}_{#2}, ..., {#1}_{#3} \rangle}
\newcommand{\sequencewithlength}[2]{\sequencewithbeginning{#1}{1}{#2}}

\newcommand{\sequencerep}[1]{\bar{#1}}
\newcommand{\modality}[1]{#1}

\newcommand{\power}[1]{^{#1}}

\newcommand{\forward}[1]{\overrightarrow{#1}}

\newcommand{\outputside}{o}

\newcommand{\allouttokenside}{y}
\newcommand{\allintokenside}{x}

\newcommand{\outputvocab}{\vocabulary\power\outputside}

\newcommand{\intokenembedding}{\embedding\power\allintokenside}
\newcommand{\outtokenembedding}{\embedding\power\allouttokenside}

\newcommand{\outputtoken}{y}

\newcommand{\snippetname}{\bar{s}}
\newcommand{\interactionname}{I}
\newcommand{\inputtoken}{x}
\newcommand{\utterancename}{\sequencerep{\inputtoken}}
\newcommand{\interactionlength}{n}
\newcommand{\interaction}{\sequencerep{\interactionname}}

\newcommand{\utteranceageemb}{\embedding^\interactionname}

\newcommand{\encoder}{^\modality{E}}
\newcommand{\hiddenstate}{\vectorrep{h}}
\newcommand{\cellmemory}{\vectorrep{c}}

\newcommand{\lstm}{\text{LSTM}}

\newcommand{\forwardutterancelstm}{\lstm^{\forward{E}}}
\newcommand{\forwardencoderstate}{\hiddenstate^{\forward{E}}}
\newcommand{\backwardencoderstate}{\hiddenstate^{\overleftarrow{E}}}

\newcommand{\encoderstate}{\hiddenstate\encoder}
\newcommand{\encodercellstate}{\cellmemory\encoder}

\newcommand{\discoursemodality}{\power{\modality{\interactionname}}}
\newcommand{\discourselstm}{\lstm\discoursemodality}
\newcommand{\discoursestate}{\hiddenstate\discoursemodality}

\newcommand{\utteranceindex}{i}
\newcommand{\inputtokenindex}{j}
\newcommand{\historylength}{h}
\newcommand{\otherutteranceindex}{t}
\newcommand{\outputtokenindex}{k}

\newcommand{\maxsnippetage}{g}
\newcommand{\prevqueryside}{^\modality{Q}}
\newcommand{\snippetside}{^\modality{S}}
\newcommand{\snippetset}{S}

\newcommand{\snippetleftidx}{l}
\newcommand{\snippetrightidx}{r}
\newcommand{\snippetage}{a}

\newcommand{\snippetencoding}{\hiddenstate\snippetside}

\newcommand{\previousquerystate}{\hiddenstate\prevqueryside}

\newcommand{\snippetageemb}{\embedding^\maxsnippetage}

\newcommand{\loss}{\mathcal{L}}

\newcommand{\decodermodality}{\power{\modality{D}}}
\newcommand{\decoder}{\lstm\decodermodality}
\newcommand{\decoderstate}{\hiddenstate\decodermodality}

\newcommand{\contextvector}{\vectorrep{c}}

\newcommand{\weightmatrix}{\vectorrep{W}}
\newcommand{\bias}{\vectorrep{b}}

\newcommand{\attentionscore}{s}
\newcommand{\attentionmodality}{\modality{A}}
\newcommand{\attentionbilinear}{\vectorrep{\weightmatrix\power\attentionmodality}}
\newcommand{\attentiondist}{\alpha}

\newcommand{\intermediatename}{m}
\newcommand{\intermediatevector}{\vectorrep{\intermediatename}}
\newcommand{\intermediateweights}{\weightmatrix\power\intermediatename}

\newcommand{\outputvocabweights}{\vectorrep{W}\power\outputside}
\newcommand{\outputvocabbias}{\bias\power\outputside}

\newcommand{\snippetweights}{\weightmatrix\power\snippetside}

\newcommand{\goldquery}{\sequencerep{\outputtoken}}

\newcommand{\stdev}[2]{${#1}$\begin{tiny}$\pm {#2}$\end{tiny}}

\newcommand{\textbox}[1]{\vspace{-0.2em}\begin{center}\fbox{\parbox{0.95\columnwidth}{#1}}\end{center}}

\newcommand{\len}[1]{{|#1|}}

\newcommand{\token}{\inputtoken}
\newcommand{\utterance}{\utterancename}
\newcommand{\allutterance}{\mathcal{X}}

\newcommand{\querytoken}{y}
\newcommand{\allquery}{\mathcal{Y}}
\newcommand{\query}{\sequencerep{\querytoken}}

\newcommand{\allinteraction}{\mathcal{I}}
\newcommand{\intind}{l}
\newcommand{\uind}{\utteranceindex}
\newcommand{\tind}{\inputtokenindex}

\author[$\dagger$]{Alane Suhr}
\author[$\ddagger$]{Srinivasan Iyer}
\author[$\dagger$]{Yoav Artzi}
\affil[$\dagger$]{
Dept. of Computer Science and Cornell Tech\\
Cornell University\\
New York, NY \authorcr
{\tt \{suhr, yoav\}@cs.cornell.edu}
}
\affil[$\ddagger$]{
Paul G. Allen School of Computer Science \& Engineering\\
Univ. of Washington\\
Seattle, WA \authorcr
{\tt sviyer@cs.washington.edu}
}

\date{}

\title{Learning to Map Context-Dependent Sentences \\ to Executable Formal Queries}

\begin{document}
\maketitle

\begin{abstract}

We propose a context-dependent model to map utterances within an interaction to executable formal queries. To incorporate interaction history, the model maintains an interaction-level encoder that updates after each turn, and can copy sub-sequences of previously predicted queries during generation. Our approach combines implicit and explicit modeling of references between utterances. We evaluate our model on the ATIS flight planning interactions, and demonstrate the benefits of modeling context and explicit references. 
\end{abstract}

\section{Introduction}
\label{sec:intro}

The meaning of conversational utterances depends strongly on the history of the interaction. 
Consider a user querying a flight database using natural language (Figure~\ref{fig:conv}). 
Given a user utterance, the system must generate a query, execute it, and display results to the user, who then provides the next request. 
Key to correctly mapping utterances to executable queries is resolving references. 
For example, the second utterance implicitly depends on the first, and the reference \nlstring{ones} in the third utterance  explicitly refers to the response to the second utterance. 
Within an interactive system, this information needs to be composed with mentions of database entries (e.g., \nlstring{Seattle}, \nlstring{next Monday}) to generate a formal executable representation. 
In this paper, we propose encoder-decoder models that directly map user utterances to executable queries, while considering the history of the interaction, including both previous utterances and their generated queries.

\begin{figure}
\fbox{
\centering
\begin{minipage}{0.95\linewidth}
	\footnotesize
	\nlstring{show me flights from seattle to boston next monday} \\[1pt]
	[Table with $31$ flights] \\[3pt]
	\nlstring{on american airlines} \\[1pt]
	[Table with $5$ flights] \\[3pt]
	\nlstring{which ones arrive at 7pm} \\[1pt]
	[No flights returned] \\[3pt]
	\nlstring{show me delta flights} \\[1pt]
	[Table with $5$ flights] \\
	$\dots$
\end{minipage}}
\caption{An excerpt of an interaction from the ATIS  flight planning system~\cite{Hemphill:90atis,Dahl:94}. Each request is followed by a description of the system response.}
\label{fig:conv}	
\end{figure}

Reasoning about how the meaning of an utterance depends on the history of the interaction is critical to correctly respond to user requests. 
As interactions progress, users may omit previously-mentioned constraints and entities, and an increasing portion of the utterance meaning must be derived from the interaction history. 
Figure~\ref{fig:convsql} shows SQL queries for the utterances in Figure~\ref{fig:conv}. 
As the interaction progresses, the majority of the generated query is derived from the interaction history (underlined), rather than from the current utterance.
A key challenge is resolving what past information is incorporated and how.
For example, in the figure, the second utterance depends on the set of flights defined by the first, while adding a new constraint. The third utterance further refines this set by adding a constraint to the constraints from both previous utterances. 
In contrast, the fourth utterance refers only to the first one, and skips the two utterances in between.\footnote{An alternative explanation is that utterance four refers to utterance three, and deletes the time and airline constraints.} 
Correctly generating the fourth query requires understanding that the time constraint (\nlstring{at 7pm}) can be ignored as it follows an airline constraint that has been replaced.

\bgroup
\newcommand{\pastturnref}[1]{\ul{#1}}
\begin{figure*}
	\centering
	\scriptsize
	\fbox{
	\begin{minipage}{0.98\textwidth}
		\hspace{-11pt}
		\begin{tabular}{p{0.02cm}p{15.34cm}}
		$\utterancename_1$: & \nlstring{show me flights from seattle to boston next monday} \\
		$\goldquery_1$: & \begin{scriptsize}\sqlstring{(SELECT DISTINCT flight.flight\_id FROM flight WHERE (flight.from\_airport IN (SELECT  \mbox{airport\_service.airport\_code} FROM airport\_service WHERE airport\_service.city\_code IN (SELECT city.city\_code FROM city WHERE city.city\_name = 'SEATTLE'))) AND (flight.to\_airport IN (SELECT airport\_service.airport\_code FROM airport\_service WHERE airport\_service.city\_code IN (SELECT city.city\_code FROM city WHERE city.city\_name = 'BOSTON'))) AND (flight.flight\_days IN (SELECT days.days\_code FROM days WHERE days.day\_name IN (SELECT date\_day.day\_name FROM date\_day WHERE date\_day.year = 1993 AND date\_day.month\_number = 2 AND date\_day.day\_number = 8))));
}\end{scriptsize} \\[2pt]
		$\utterancename_2$: & \nlstring{on american airlines} \\
		$\goldquery_2$: & \begin{scriptsize}\sqlstring{(SELECT \pastturnref{DISTINCT flight.flight\_id FROM flight} WHERE (flight.airline\_code = 'AA') AND \ul{(flight.from\_airport IN (SELECT airport\_service.airport\_code FROM airport\_service WHERE airport\_service.city\_code IN (SELECT city.city\_code FROM city WHERE city.city\_name = 'SEATTLE')))} AND \ul{(flight.to\_airport IN (SELECT airport\_service.airport\_code FROM airport\_service WHERE airport\_service.city\_code IN (SELECT city.city\_code FROM city WHERE city.city\_name = 'BOSTON')))} AND \ul{(flight.flight\_days IN (SELECT days.days\_code FROM days WHERE days.day\_name IN (SELECT date\_day.day\_name FROM date\_day WHERE date\_day.year = 1993 AND date\_day.month\_number = 2 AND date\_day.day\_number = 8)))});
		}\end{scriptsize} \\[2pt]
		$\utterancename_3$: & \nlstring{which ones arrive at 7pm} \\
		$\goldquery_3$: & \begin{scriptsize}\sqlstring{(SELECT \pastturnref{DISTINCT flight.flight\_id FROM flight} WHERE \ul{(flight.airline\_code = 'AA')} AND \ul{(flight.from\_airport IN (SELECT airport\_service.airport\_code FROM airport\_service WHERE airport\_service.city\_code IN (SELECT city.city\_code FROM city WHERE city.city\_name = 'SEATTLE')))} AND \ul{(flight.to\_airport IN (SELECT airport\_service.airport\_code FROM airport\_service WHERE airport\_service.city\_code IN (SELECT city.city\_code FROM city WHERE city.city\_name = 'BOSTON')))} AND \ul{(flight.flight\_days IN (SELECT days.days\_code FROM days WHERE days.day\_name IN (SELECT date\_day.day\_name FROM date\_day WHERE date\_day.year = 1993 AND date\_day.month\_number = 2 AND date\_day.day\_number = 8)))} AND (flight.arrival\_time = 1900));
}\end{scriptsize} \\[2pt]
		$\utterancename_4$: & \nlstring{show me delta flights} \\
		$\goldquery_4$: & \begin{scriptsize}\sqlstring{(SELECT DISTINCT flight.flight\_id FROM flight WHERE (flight.airline\_code = 'DL') AND \ul{(flight.from\_airport IN (SELECT airport\_service.airport\_code FROM airport\_service WHERE airport\_service.city\_code IN (SELECT city.city\_code FROM city WHERE city.city\_name = 'SEATTLE')))} AND \ul{(flight.to\_airport IN (SELECT airport\_service.airport\_code FROM airport\_service WHERE airport\_service.city\_code IN (SELECT city.city\_code FROM city WHERE city.city\_name = 'BOSTON')))} AND \ul{(flight.flight\_days IN (SELECT days.days\_code FROM days WHERE days.day\_name IN (SELECT date\_day.day\_name FROM date\_day WHERE date\_day.year = 1993 AND date\_day.month\_number = 2 AND date\_day.day\_number = 8)))});
		}\end{scriptsize}
		\end{tabular}
	\end{minipage}}
	\caption{Annotated SQL queries ($\goldquery_1$,\dots,$\goldquery_4$) in the ATIS~\cite{Hemphill:90atis} domain for utterances ($\utterancename_1$,\dots,$\utterancename_4$) from Figure~\ref{fig:conv}. Underlining (not part of the annotation) indicates segments originating from the interaction context.}
	\label{fig:convsql}
\end{figure*}
\egroup

We study complementary methods to enable this type of reasoning. 
The first set of methods implicitly reason about references by modifying the encoder-decoder architecture to encode information from previous utterances for generation decisions. 
We experiment with attending over previous utterances and using an interaction-level recurrent encoder. 
We also study explicitly maintaining a set of referents using segments from previous queries. 
At each step, the decoder chooses whether to output a token or select a segment from the set, which is appended to the output in a single decoding step.  
In addition to enabling references to previously mentioned entities, sets, and constraints, this method also reduces the number of generation steps required, illustrated by the underlined segments in Figure~\ref{fig:convsql}. 
For example, the query $\goldquery_2$ will require $17$ steps instead of $94$.

We evaluate our approach using the ATIS~\cite{Hemphill:90atis,Dahl:94} task, where a user interacts with a SQL flight database using natural language requests, and almost all queries require joins across multiple tables. 
In addition to reasoning about contextual phenomena, we design our system to effectively resolve database values, including resolution of time expressions (e.g., \nlstring{next monday} in Figure~\ref{fig:conv}) using an existing semantic parser. 
Our evaluation shows that reasoning about the history of the interaction is necessary, relatively increasing performance by $28.6\%$  over a baseline with no access to this information, and that combining the implicit and explicit methods provides the best performance.  
Furthermore, our analysis shows that our full approach maintains its performance as interaction length increases, while the performance of systems without explicit modeling deteriorates.
Our code is available at \url{https://github.com/clic-lab/atis}.

\section{Technical Overview}
\label{sec:overview}

Our goal is to map utterances in interactions to formal executable queries. 
We evaluate our approach with the ATIS corpus~\cite{Hemphill:90atis,Dahl:94}, where users query a realistic flight planning system using natural language. The system responds by displaying tables and database entries. 
User utterances are mapped to SQL to query a complex database with $27$ tables and $162$K entries. $96.6\%$ of the queries require joins of different tables. 
Section~\ref{sec:exp} describes ATIS.

\paragraph{Task Notation}

Let $\allinteraction$ be the set of all interactions, $\allutterance$ the set of all utterances, and $\allquery$ the set of all formal queries. 
A user utterance $\utterance \in \allutterance$ of length $\len{\utterance}$ is a sequence $\langle \token_1, \dots, \token_{\len{\utterance}}\rangle$, where each $\token_i$ is a natural language token.
A formal query $\query\in\allquery$ of length $\len{\query}$ is a sequence $\langle \querytoken_1, \dots, \querytoken_{\len{\query}}\rangle$, where each $\querytoken_i$ is a formal query token. 
An interaction $\interaction \in \allinteraction$ is a sequence of $\interactionlength$ utterance-query pairs $\langle(\utterance_1, \query_1),\dots,(\utterance_\interactionlength, \query_\interactionlength)\rangle$ representing an interaction with $\interactionlength$ turns. 
To refer to indexed interactions and their content, we mark $\interaction^{(\intind)}$ as an interaction with index $\intind$, the $\uind$-th utterance and query in $\interaction^{(\intind)}$ as $\utterance^{(\intind)}_{\uind}$ and $\query^{(\intind)}_{\uind}$, and the $\tind$-th tokens in $\utterance^{(\intind)}_{\uind}$ and $\query^{(\intind)}_{\uind}$ as $\token^{(\intind)}_{\uind,\tind}$ and $\querytoken^{(\intind)}_{\uind,\tind}$. 
At turn $\uind$, we denote the interaction history of length $\uind-1$ as $\interaction[:\uind-1] = \langle(\utterance_1, \query_1),\dots,(\utterance_{\uind-1}, \query_{\uind-1})\rangle$.
Given $\interaction[:{\uind-1}]$ and utterance $\utterance_{\uind}$ our  goal is to generate $\query_{\uind}$, while considering both $\utterance_{\uind}$ and $\interaction[:\uind-1]$.
Following the execution of $\query_{\uind}$, the interaction history at turn $\uind+1$ becomes $\interaction[:\uind] = \langle(\utterance_1, \query_1),\dots,(\utterance_{\uind}, \query_{\uind})\rangle$.

\paragraph{Model}

Our model is based on the recurrent neural network~\cite[RNN;][]{Elman:90rnn} encoder-decoder framework with attention~\cite{Cho:14nmt,Sutskever:14,Bahdanau:14neuralmt,Luong:15nmtattention}. 
We modify the model in three ways to reason about context from the interaction history by attending over previous utterances (Section~\ref{sec:multiple_utterance_attention}), adding a turn-level recurrent encoder that updates after each turn (Section~\ref{sec:discourse_lstm}), and adding a mechanism to copy segments of queries from previous utterances (Section~\ref{sec:snippets}). 
We also design a scoring function to score values that are abstracted during pre-processing, including entities and times (Section~\ref{sec:anon}).
The full model selects between generating query tokens and copying complete segments from previous queries.

\paragraph{Learning}
We assume access to a training set that contains $N$ interactions $\{\interaction^{(\intind)}\}_{ \intind = 1}^N$. 
We train using a token-level cross-entropy objective (Section~\ref{sec:learning}). 
For models that use the turn-level encoder, we construct computational graphs for the entire interaction and back-propagate the loss for all queries together. 
Without the turn-level encoder, each utterance is processed separately.

\paragraph{Evaluation}

We evaluate using a test set $\{\interaction^{(\intind)}\}_{ \intind = 1}^M$ of $M$ interactions. 
We measure the accuracy of each utterance for each test interaction against the annotated query and its execution result. 
For models that copy segments from previous queries, we evaluate using both predicted and gold previous queries.

\section{Related Work}
\label{sec:related}

Mapping sentences to formal representations, commonly known as semantic parsing, has been studied extensively with linguistically-motivated compositional representations, including variable-free logic~\cite[e.g.,][]{Zelle:96results,Clarke:10}, lambda calculus~\cite[e.g.,][]{Zettlemoyer:05,Artzi:11,Kushman:13}, and dependency-based compositional semantics~\citep[e.g.,][]{Liang:11,Berant:13}. 
Recovering lambda-calculus representations was also studied with ATIS with focus on context-independent meaning using grammar-based approaches~\cite{Zettlemoyer:07,Kwiatkowski:11,Wang:14atis} and neural networks~\cite{Dong16:lang-to-logic,Jia:16recombination}.

Recovering context-independent executable representations has been receiving increasing attention. 
Mapping sentence in isolation to SQL queries has been studied with ATIS using statistical parsing~\cite{Popescu:04atis,Poon:13gusp} and sequence-to-sequence models~\cite{Iyer:16sempar-feedback}. 
Generating executable programs was studied with other domains and formal languages~\cite{Giordani:12,Ling:16,Zhong:17,Xu:17}. 
Recently, various approaches were proposed to use the formal language syntax  to constrain the search space~\cite{Yin:17syntax-code,Rabinovich:17syntax-code,Krishnamurthy:17,Cheng:17} making all outputs valid programs. 
These contributions are orthogonal to ours, and can be directly integrated into our decoder. 

Generating context-dependent formal representations has received less attention. 
\citet{Miller:96} used ATIS and mapped utterances to semantic frames, which were then mapped to SQL queries.
For learning, they required full supervision, including annotated parse trees and contextual dependencies.\footnote{\citet{Miller:96} provide limited details about their evaluation. Later work notes that they evaluate SQL query correctness~\cite{Zettlemoyer:09} with an accuracy of $78.4\%$, higher than our results. However, the lack of details (e.g., if the metric is strict or relaxed) makes comparison difficult. In addition, we use significantly less supervision, and re-split the data to avoid scenario bias (Section~\ref{sec:exp}).} 
\citet{Zettlemoyer:09} addressed the problem with lambda calculus, using a semantic parser trained separately with context-independent data. 
In contrast, we generate executable formal queries and require only interaction query annotations for training.

Recovering context-dependent meaning was also studied with the SCONE~\cite{Long:16context} and SequentialQA~\cite{Iyyer:17seq-qa} corpora.
We compare ATIS to these corpora in Section~\ref{sec:exp}. 
Resolving explicit references, a part of our problem, has been studied as co-reference resolution~\cite{Ng:10coref}.
Context-dependent language understanding was also studied for dialogue systems, including with ATIS, as surveyed by~\citet{Tur:10atis}. 
More recently, encoder-decoder methods were applied to dialogue systems~\cite{Peng:17comp-task-dialog,Li:17dialogue}, including using hierarchical RNNs~\cite{Serban:16,Serban:17}, an architecture related to our turn-level encoder. 
These approaches use slot-filling frames with limited expressivity, while we focus on the original representation of unconstrained SQL queries.

\section{Context-dependent Model}
\label{sec:models}

\begin{figure*}
\begin{center}
\includegraphics[width=0.97\textwidth,trim={59pt 244pt 44pt 127pt},clip=true]{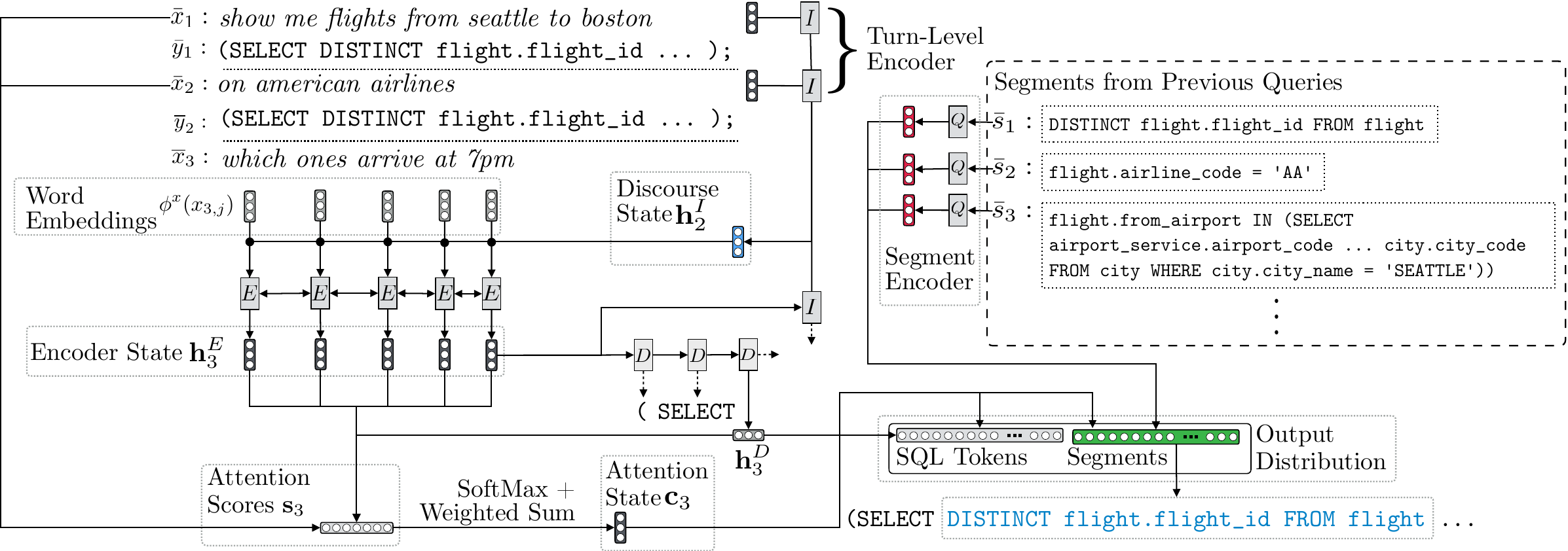}
\caption{
Illustration of the model architecture during the third decoding step while processing the instruction \nlstring{which ones arrive at 7pm} from the interaction in Figure~\ref{fig:convsql}.
The current discourse state $\discoursestate_2$ is used to encode the current utterance $\utterancename_3$ (Section~\ref{sec:discourse_lstm}).
Query segments from previous queries are encoded into vector representations (Section~\ref{sec:snippets}).
In each generation step, the decoder attends over the previous and current utterances, and a probability distribution is computed over SQL tokens and query segments. Here, segment $\snippetname_1$ is selected. }
\label{fig:model_diagram}
\end{center}
\end{figure*}

We base our model on an encoder-decoder architecture with attention~\cite{Cho:14nmt,Sutskever:14,Bahdanau:14neuralmt,Luong:15nmtattention}.
At each interaction turn $\utteranceindex$, given the current utterance $\utterance_\utteranceindex$ and the interaction history $\interaction[:\utteranceindex-1]$, the model generates the formal query $\query_i$. 
Figure~\ref{fig:model_diagram} illustrates our architecture. 
We describe the base architecture, and gradually add components. 

\subsection{Base Encoder-Decoder Architecture}
\label{sec:base_model}

Our base architecture uses an encoder to process the user utterance $\utterance_\utteranceindex = \langle \token_{\utteranceindex,1},\dots,\token_{\utteranceindex,\len{\utterance_\utteranceindex}} \rangle$ and a decoder to generate the output query $\query_i$ token-by-token. 
This architecture does not observe the interaction history $\interaction[:\utteranceindex-1]$.

The  encoder computes a hidden state $\encoderstate_\tind = [\forwardencoderstate_{\inputtokenindex} ; \backwardencoderstate_{\inputtokenindex}]$ for each token $\token_{\uind,\tind}$ using a bi-directional RNN. The forward RNN is defined by:~\footnote{We omit the memory cell (often denoted as $\mathbf{c}_\inputtokenindex$) from all LSTM descriptions. We use only the LSTM hidden state $\mathbf{h}_\inputtokenindex$ in other parts of the architecture unless explicitly noted.}

\begin{small}
\begin{equation}
	\label{eq:plain_encoder} \forwardencoderstate_{\inputtokenindex}  = \forwardutterancelstm\left( \intokenembedding(\inputtoken_{\utteranceindex, \inputtokenindex}) ; \forwardencoderstate_{\inputtokenindex-1} \right)\;\;,
\end{equation}
\end{small}

\noindent
where $\forwardutterancelstm$ is a long short-term memory recurrence~\cite[LSTM;][]{Hochreiter:97lstm} and $\intokenembedding$ is a learned embedding function for input tokens.
The backward RNN recurs  in the opposite direction with separate parameters. 

We generate the query with an RNN decoder. The decoder state at step $\outputtokenindex$ is:  

\begin{small}
\begin{equation*}
	\decoderstate_\outputtokenindex = \decoder\left( \left[ \outtokenembedding(\outputtoken_{\uind,\outputtokenindex-1}); \contextvector_{\outputtokenindex-1} \right]; \decoderstate_{\outputtokenindex-1}\right)\;\;,
\end{equation*}
\end{small}

\noindent
where $\decoder$ is a two-layer LSTM recurrence,  $\outtokenembedding$ is a learned embedding function for query tokens, and $\contextvector_\outputtokenindex$ is an attention vector computed from the encoder states. $\outputtoken_{\utteranceindex, 0}$ is a special start token, and $\contextvector_{0}$ is a zero-vector. 
The initial hidden state and cell memory of each layer are initialized as $\encoderstate_{\length{\utterancename_\utteranceindex}}$ and $\encodercellstate_{\length{\utterancename_\utteranceindex}}$.
The attention vector $\contextvector_\outputtokenindex$ is a weighted sum of the encoder hidden states: 

\begin{small}
\begin{eqnarray}
\label{eq:attention} 				\attentionscore_{\outputtokenindex}(\inputtokenindex) &=& \encoderstate_{\inputtokenindex} \attentionbilinear \decoderstate_\outputtokenindex \\	
\label{eq:attention_distribution}	\attentiondist_{\outputtokenindex} &=& \text{softmax}(\vectorrep{\attentionscore}_{\outputtokenindex}) \\
\label{eq:context_vector}			\contextvector_{\outputtokenindex} &=& \sum_{\inputtokenindex=1}^{\length{\utterancename_\utteranceindex}} \encoderstate_{\inputtokenindex} \attentiondist_{\outputtokenindex}(\inputtokenindex)\;\;,
\end{eqnarray}
\end{small}

\noindent
where $\attentionbilinear$ is a learned matrix. 
The probabilities of output query tokens are computed as: 

\begin{small}
\begin{eqnarray}
\label{eq:intermediate_vector}	\intermediatevector_{\outputtokenindex} &=& \tanh\left( [ \decoderstate_{\outputtokenindex}; \contextvector_{\outputtokenindex} ] \intermediateweights \right)~~ \\
\label{eq:plain_probdist} 	P(\outputtoken_{\uind,\outputtokenindex} = w \mid \utterance_\uind, \query_{\uind,1:\outputtokenindex-1}  ) &\propto & \exp(\intermediatevector_{\outputtokenindex}\outputvocabweights_w  + \outputvocabbias_w)
\end{eqnarray}
\end{small}

\noindent
where $\intermediateweights$, $\outputvocabweights$, and $\outputvocabbias$ are learned.

\subsection{Incorporating Recent History}
\label{sec:multiple_utterance_attention}

We provide the model with the most recent interaction history by concatenating the previous $\historylength$ utterances $\sequencewithbeginning{\utterancename}{\utteranceindex-\historylength}{\utteranceindex-1}$ with the current utterance in order, adding a special delimiter token between each utterance.
The concatenated input provides the model access to previous utterances, but not to previously generated queries, or utterances that are more than $\historylength$ turns in the past. 
The architecture remains the same, except that the encoder and attention are computed over the concatenated sequence of tokens. 
The probability of an output query token is computed the same, but is now conditioned on the interaction history:

\begin{small}
\begin{eqnarray}
\label{eq:concat_prob} P(\outputtoken_{\uind,\outputtokenindex} = w \mid \utterance_\uind, \query_{\uind,1:\outputtokenindex-1}, \interaction[:\utteranceindex-1] ) \propto && \\ \nonumber && \hspace{-2cm} \exp(\intermediatevector_{\outputtokenindex}\outputvocabweights_w  + \outputvocabbias_w)\;\;.
\end{eqnarray}
\end{small}

\subsection{Turn-level Encoder}
\label{sec:discourse_lstm}

Concatenating recent utterances to provide access to recent history has computational drawbacks. 
The encoding of the utterance depends on its location in the concatenated string. 
This requires encoding all recent history for each new utterance, and does not allow re-use of computation between utterances during encoding. 
It also introduces a tradeoff between computation cost and expressivity: attending over the $\historylength$ previous utterances allows the decoder access to the information in these utterances when generating a query, but is computationally more expensive as $\historylength$ increases. 
We address this by encoding each utterance once. 
To account for the influence of the interaction history on utterance encoding, we maintain a discourse state encoding $\discoursestate_{\utteranceindex}$ computed with a turn-level recurrence, and use it during utterance encoding. 
The state is maintained and updated over the entire interaction. 
At turn $\utteranceindex$, this model has access to the complete prefix of the interaction $\interaction[:\utteranceindex-1]$ and the current request $\utterancename_\utteranceindex$. 
In contrast, the concatenation-based encoder (Section~\ref{sec:multiple_utterance_attention})  has access only to information from the previous $\historylength$ utterances. 
We also use positional encoding in the attention computation to account for the position of each utterance relative to the current utterance.

Formally, we modify Equation~\ref{eq:plain_encoder} to encode $\utterancename_\utteranceindex$:

\begin{small}
\begin{equation*}
\nonumber	\forwardencoderstate_{\utteranceindex, \inputtokenindex} = \forwardutterancelstm\left( \left[ \intokenembedding(\inputtoken_{\utteranceindex, \inputtokenindex}) ; \discoursestate_{\utteranceindex-1} \right]; \forwardencoderstate_{\utteranceindex, \inputtokenindex-1} \right)\;\;,
\end{equation*}
\end{small}

\noindent
 where $\discoursestate_{\utteranceindex-1}$ is the discourse state following utterance $\utterancename_{\utteranceindex-1}$. 
$\text{LSTM}^{\overleftarrow{E}}$ is modified analogously. 
In contrast to the concatenation-based model, the recurrence processes a single utterance.
The discourse state $\discoursestate_{\utteranceindex}$ is computed as 

\begin{small}
\begin{equation}
\nonumber \discoursestate_{\utteranceindex} = \discourselstm\left( \encoderstate_{\utteranceindex,\length{\utterancename_{\utteranceindex}}}; \discoursestate_{\utteranceindex-1}\right)\;\;.
\end{equation}
\end{small}

\noindent
Similar to the concatenation-based model, we attend over the current utterance and the $\historylength$ previous utterances. 
We add relative position embeddings $\utteranceageemb$ to each hidden state. These embeddings are learned for each possible distance $0,\dots,\historylength-1$ from the current utterance. 
We modify Equation~\ref{eq:attention} to index over both utterances and tokens:

\begin{small}
\begin{equation}\label{eq:turn_level_attention}
	\attentionscore_{\outputtokenindex}(\otherutteranceindex, \inputtokenindex) = \left[\encoderstate_{\otherutteranceindex, \inputtokenindex}; \utteranceageemb(\utteranceindex-\otherutteranceindex) \right] \attentionbilinear \decoderstate_\outputtokenindex\;\;.
\end{equation}
\end{small}

\noindent
In contrast to the concatenation model, without position embeddings, the attention computation has no indication of the utterance position, as our ablation shows in Section~\ref{sec:results}. 
The attention distribution is computed as in Equation~\ref{eq:attention_distribution}, and normalized across all utterances. 
The position embedding is also used to compute the context vector $\contextvector_{\outputtokenindex}$:

\begin{small}
\begin{equation*}
\contextvector_{\outputtokenindex} = \sum_{\otherutteranceindex=\utteranceindex-\historylength}^{\utteranceindex} \sum_{\inputtokenindex=1}^{\length{\utterancename_{\otherutteranceindex}}} \left[ \encoderstate_{\otherutteranceindex, \inputtokenindex} ; \utteranceageemb(\utteranceindex-\otherutteranceindex)\right] \attentiondist_{\outputtokenindex}(\otherutteranceindex, \inputtokenindex)\;\;.
\end{equation*}
\end{small}

\subsection{Copying Query Segments}
\label{sec:snippets}

The discourse state and attention over previous utterances allow the model to consider the interaction history when generating queries. 
However, we observe that context-dependent reasoning often requires generating sequences that were generated in previous turns. 
Figure~\ref{fig:convsql} shows how segments (underlined) extracted from previous utterances are predominant in later queries. 
To take advantage of what was previously generated, we add copying of complete segments from previous queries by expanding the set of outputs at each generation step.  
This mechanism explicitly models references, reduces the number of steps required to generate queries, and provides an interpretable view of what parts of a query originate in context. 
Figure~\ref{fig:model_diagram} illustrates this architecture.

\paragraph{Extracting Segments}
Given the interaction history $\interaction[:\utteranceindex-1]$, we construct the set of segments $\snippetset_{\utteranceindex-1}$ by deterministically extracting sub-trees from previously generated queries.\footnote{The process of extracting sub-trees is described in the supplementary material.}  
In our data, we extract $13\pm 5.9$ ($\mu\pm\sigma$) segments for each annotated query. 
Each segment $\snippetname \in \snippetset_{\utteranceindex-1}$ is a tuple $\langle\snippetage, b, \snippetleftidx, \snippetrightidx \rangle$, where $\snippetage$ and $b$ are the indices of the first and most recent queries, $\query_\snippetage$ and $\query_b$, in the interaction that contain the segment. $\snippetleftidx$ and $\snippetrightidx$ are the start and end indices of the segment in $\query_b$.

\paragraph{Encoding Segments}

We represent a segment $\snippetname =  \langle \snippetage, b, \snippetleftidx, \snippetrightidx \rangle$ using the hidden states of an RNN encoding of the query $\query_b$. 
The hidden states $\sequencewithlength{\previousquerystate}{\length{\query_b}}$  are computed using a bi-directional LSTM RNN similar to the utterance encoder (Equation~\ref{eq:plain_encoder}), except using separate LSTM parameters and $\outtokenembedding$ to embed the query tokens. 
The embedded representation of a segment is a concatenation of the hidden states at the segment endpoints and an embedding of the relative position of the utterance where it appears first:

\begin{small}
\begin{equation*}
	\snippetencoding = \left[ \previousquerystate_{\snippetleftidx}; \previousquerystate_{\snippetrightidx};  \snippetageemb(\min(\maxsnippetage, i-\snippetage)) \right]\;\;,
\end{equation*}
\end{small}

\noindent
where $\snippetageemb$ is a learned embedding function of the position of the initial query $\query_\snippetage$ relative to the current turn index $\utteranceindex$. 
We learn an embedding for each relative position that is smaller than $\maxsnippetage$, and use the same embedding for all other positions. 

\paragraph{Generation with Segments}

At each generation step, the decoder selects between a single query token or a segment. 
When a segment is selected, it is appended to the generated query, an embedded segment representation for the next step is computed, and generation continues. 
The probability of a segment $\snippetname =  \langle \snippetage, b, \snippetleftidx, \snippetrightidx \rangle$ at decoding step $\outputtokenindex$ is:

\begin{small}
\begin{eqnarray}
\label{eq:probdist_with_snippets}	P(\outputtoken_{\utteranceindex,\outputtokenindex} = \snippetname \mid \utterancename_\utteranceindex, \query_{\utteranceindex,1:\outputtokenindex -1}, \interaction[:\utteranceindex-1] ) \propto && \\ \nonumber && \hspace{-50pt} \exp \left( \intermediatevector_\outputtokenindex \snippetweights \snippetencoding \right)\;\;,
\end{eqnarray}
\end{small}

\noindent
where $\intermediatevector_\outputtokenindex$ is computed in Equation~\ref{eq:intermediate_vector} and $\snippetweights$ is a learned matrix. 
To simplify the notation, we assign the segment to a single output token. 
The output probabilities (Equations~\ref{eq:concat_prob} and~\ref{eq:probdist_with_snippets}) are normalized together to a single probability distribution. 
When a segment is selected, the  embedding used as input for the next generation step is a bag-of-words encoding of the segment. We extend the output token function $\outtokenembedding$ to take segments:

\begin{small}
\begin{equation*}
\outtokenembedding(\snippetname = \langle \snippetage, b, \snippetleftidx, \snippetrightidx\rangle) = \dfrac{1}{\snippetrightidx-\snippetleftidx} \sum_{k=\snippetleftidx}^{\snippetrightidx} \outtokenembedding \left( \outputtoken_{b, k}\right)\;\;.
\end{equation*}
\end{small}

\noindent
The recursion in $\outtokenembedding$ is limited to depth one because segments do not contain other segments.

\subsection{Inference with Full Model}
\label{sec:model:inference}

Given an utterance $\utterancename_\utteranceindex$ and the history of interaction $\interaction[:\uind-1]$, we generate the query $\query_\utteranceindex$. 
An interaction starts with the user providing the first utterance $\utterancename_1$. The utterance is encoded using the initial discourse state $\discoursestate_0$, the discourse state $\discoursestate_1$ is computed, the query $\query_1$ is generated, and the set of segments $\snippetset_1$ is created. 
The initial discourse state $\discoursestate_0$ is learned, and the set of segments $\snippetset_0$ used when generating $\query_1$ is the empty set. 
The attention is computed only over the first utterance because no previous utterances exist. 
The user then provides the next utterance or concludes the interaction. 
At turn $\utteranceindex$, the utterance $\utterance_\utteranceindex$ is encoded using the discourse state $\discoursestate_{\utteranceindex-1}$, the discourse state $\discoursestate_{\utteranceindex}$ is computed, and the query $\query_\utteranceindex$ is generated using the set of segments $\snippetset_{\utteranceindex-1}$. 
The model has no access to future utterances. We use greedy inference for generation. 
Figure~\ref{fig:model_diagram} illustrates a single decoding step.

\section{Learning}
\label{sec:learning}

We assume access to a training set of $N$ interactions $\{\interaction^{(\intind)}\}_{ \intind = 1}^N$. 
Given an interaction $\interaction^{(\intind)}$, each utterance $\utterance^{(\intind)}_\utteranceindex$ where $1 \leq \utteranceindex \leq \len{\interaction^{(\intind)}}$, is paired with an annotated query $\query^{(\intind)}_\utteranceindex$. 
The set of segments from previous utterances is deterministically extracted from the annotated queries during learning. However, the data does not indicate what parts of each query originate in segments copied from previous utterances. 
We adopt a simple approach and heuristically identify context-dependent segments based on entities that appear in the utterance and the query.\footnote{The alignment is detailed in the supplementary material.} 
Once we identify a segment in the annotated query, we replace it with a unique placeholder token, and it appears to the learning algorithm as a single generation decision.
Treating this decision as latent is an important direction for future work. 
Given the segment copy decisions, we minimize the token cross-entropy loss:

\begin{small}
\begin{equation*}
	\loss(\outputtoken^{(\intind)}_{\utteranceindex, \outputtokenindex}) = -\log P \left( \outputtoken^{(\intind)}_{\utteranceindex, \outputtokenindex} \mid \utterance^{(\intind)}_{\utteranceindex},  \query^{(\intind)}_{\utteranceindex, 1:\outputtokenindex-1}, \interaction^{(\intind)}[:\utteranceindex-1]\right)\;\;,
\end{equation*}
\end{small}

\noindent
where $\outputtokenindex$ is the index of the output token. 
The base and recent-history encoders (Sections~\ref{sec:base_model} and~\ref{sec:multiple_utterance_attention}) can be trained by processing each utterance separately. 
For these models, given a mini-batch $\mathcal{B}$ of utterances, each identified by an interaction-utterance index pair, the loss is the mean token loss

\begin{small}
\begin{equation*}
\loss = \frac{1}{\sum_{(i,j)\in \mathcal{B} } \length{\query^{(j)}_{\utteranceindex}}} \sum_{(i,j)\in \mathcal{B} }  \sum_{\outputtokenindex = 1}^{\length{\query^{(j)}_{\utteranceindex}}} \loss(\outputtoken^{(j)}_{\utteranceindex, \outputtokenindex})\;\;.
\end{equation*}
\end{small}

The turn-level encoder (Section~\ref{sec:discourse_lstm}) requires building a computation graph for the entire interaction. 
We update the model parameters for each interaction. The interaction loss is 

\begin{small}
\begin{equation*}
	\loss = \dfrac{n}{B}\frac{1}{\sum_{\utteranceindex =1}^\interactionlength \length{\query^{(j)}_{\utteranceindex}} }\sum_{\utteranceindex=1}^{\interactionlength}\sum_{\outputtokenindex=1}^{\length{\query^{(j)}_{\utteranceindex}}}\loss(\outputtoken^{(j)}_{\utteranceindex, \outputtokenindex})\;\;,
\end{equation*}	
\end{small}

\noindent
where $B$ is the batch size, and $\frac{n}{B}$ re-normalizes the loss so the gradient magnitude  is not dependent on the number of utterances in the interaction. Our ablations ($-$batch re-weight in Table~\ref{tab:results}) shows the importance of this term. 
For both cases, we use teacher forcing~\cite{Williams:89teacherforcing}.

\section{Reasoning with Anonymized Tokens}
\label{sec:anon}

An important practical consideration for generation in  ATIS and other database domains is reasoning about database values, such as entities, times, and dates. 
For example, the first utterance in Figure~\ref{fig:convsql} includes two entities and a date reference. 
With limited data, learning to both reason about a large number of entities and to resolve dates are challenging for neural network models. 
Following previous work~\cite{Dong16:lang-to-logic,Iyer:16sempar-feedback}, we address this with anonymization, where the data is pre- and post-processed to abstract over tokens that can be heuristically resolved to tokens in the query language.  
In contrast to previous work, we design a special scoring function to anonymized tokens to reflect how they are used in the input utterances. 
Figure~\ref{fig:pre-processing_example} illustrates pre-processing in ATIS. 
For example, we use a temporal semantic parser to resolve dates (e.g., \nlstring{next Monday}) and replace them with day, month, and year placeholders. 
To anonymize database entries, we use a dictionary compiled from the database (e.g., to map \nlstring{Seattle} to \texttt{SEATTLE}). The full details of the anonymization procedure are provided in the supplementary material. 
Following pre-processing, the model reasons about encoding and generation of anonymized tokens (e.g., \texttt{CITY\#1}) in addition to regular output tokens and query segments from the interaction history. 
Anonymized tokens are typed (e.g., \texttt{CITY}), map to a token in the query language (e.g., \texttt{'BOSTON'}), and appear both in input utterances and generated queries.

\begin{figure}[t]
\centering\scriptsize
\fbox{\begin{minipage}{0.95\columnwidth}
\begin{flushleft}
\ul{Original utterance and query:} \\
\hspace*{-8pt}
\begin{tabular}{p{0.02cm}p{7cm}}
$\utterancename_1$: & \nlstring{show me flights from seattle to boston next monday}  \\
$\query_1$: & \sqlstring{( SELECT DISTINCT flight.flight\_id ... city.city\_name = 'SEATTLE' ... city.city\_name = 'BOSTON' ... date\_day.year = 1993 AND date\_day.month\_number = 2 AND date\_day.day\_number = 8 ...}
\end{tabular} 
\ul{Anonymized utterance and query:} \\
\hspace*{-8pt}
\begin{tabular}{p{0.02cm}p{7cm}}
$\utterancename_1'$: & \nlstring{show me flights from CITY\#1 to CITY\#2 DAY\#1 MONTH\#1 YEAR\#1} \\
$\query_1'$: &\sqlstring{( SELECT DISTINCT flight.flight\_id ... city.city\_name = CITY\#1 ... city.city\_name = CITY\#2 ... date\_day.year = YEAR\#1 AND date\_day.month\_number = MONTH\#1 AND date\_day.day\_number = DAY\#1 ...}
\end{tabular}
\end{flushleft}
\ul{Anonymization mapping:} \\
\begin{tabular}{p{1cm}p{1.5cm}|p{1cm}p{2.5cm}}
\texttt{CITY\#1} & \texttt{'SEATTLE'} & \texttt{MONTH\#1} & \texttt{2}\\
\texttt{CITY\#2} & \texttt{'BOSTON'} & \texttt{YEAR\#1} & \texttt{1993}\\
\texttt{DAY\#1} & \texttt{8} 
\end{tabular}
\end{minipage}}
\caption{An example of date and entity anonymization pre-processing for $\utterancename_1$ and $\query_1$ in Figure~\ref{fig:convsql}.}
\label{fig:pre-processing_example}
\end{figure}

We modify our encoder and decoder embedding functions ($\intokenembedding$ and $\outtokenembedding$) to map  anonymized tokens to the embeddings of their types (e.g., \texttt{CITY}). 
The type embeddings in $\intokenembedding$ and $\outtokenembedding$  are separate. 
Using the types only, while ignoring the indices, avoids learning biases that arise from the arbitrary ordering of the tokens in the training data. 
However, it does not allow distinguishing between entries with the same type for generation decisions; for example, the common case where multiple cities are mentioned in an interaction. 
We address this by scoring anonymized token based on the magnitude of attention assigned to them at generation step $\outputtokenindex$. 
The attention magnitude is computed from the encoder hidden states. 
This computation considers both the decoder state and the location of the anonymized tokens in the input utterances to account for how they are used in the interaction.
The probability of an anonymized token $w$ at generation step $\outputtokenindex$ is

\begin{small}
\begin{eqnarray*}
	P(\querytoken_{\utteranceindex,\outputtokenindex}=w \mid \utterancename_\utteranceindex, \query_{\utteranceindex,1:\outputtokenindex -1}, \interaction[:\utteranceindex-1] ) \propto && \\ && \hspace{-60pt} \sum_{t = \utteranceindex-\historylength}^\utteranceindex \sum_{j = 1}^{\len{\utterancename_t}} \left( \exp \left( \attentionscore_\outputtokenindex \left(t,j \right) \right) \right)
\end{eqnarray*}	
\end{small}

\noindent
where $\attentionscore_\outputtokenindex \left(t,j \right)$ is the attention score computed in Equation~\ref{eq:turn_level_attention}. 
This probability is normalized together with the probabilities in Equations~\ref{eq:concat_prob} and~\ref{eq:probdist_with_snippets} to form the complete output probability.

\begin{table}[t]
\begin{footnotesize}
\begin{center}
\begin{tabular}{|l|c|} 
	\hline
	Mean/max utterances per interaction & $7.0$ /  $64$ \\ \dline
	Mean/max tokens per utterance & $10.2$ / $47$  \\ \dline
	Mean/max token per SQL query & $102.9$ / $1286$  \\ \dline
	Input vocabulary size & $1582$  \\ \dline
	Output vocabulary size & $982$  \\ \hline
	\end{tabular}
\end{center}
\end{footnotesize}
\caption{ATIS data statistics.}
\label{tab:data_stats}
\end{table}

\section{Experimental Setup}
\label{sec:exp}

Hyperparameters, architecture details, and other experimental choices  are detailed in the supplementary material.

\paragraph{Data}

We use ATIS~\cite{Hemphill:90atis,Dahl:94} to evaluate our approach.  
The data was originally collected using wizard-of-oz experiments, and annotated with SQL queries. 
Each interaction was based on a scenario given to a user. 
We observed that the original data split shares scenarios between the train, development, and test splits. 
This introduces biases, where travel patterns that appeared during training repeat in testing. 
For example, a model trained on the original data split often correctly resolves the exact referenced by \nlstring{on Saturday} with no pre-processing or access to the document date.
We evaluate this overfitting empirically in the supplementary material. 
We re-split the data to avoid this bias. 
We evenly distribute scenarios across splits so that each split contains both scenarios with many and few representative interactions. 
The new split follows the original split sizes with $1148$/$380$/$130$ train/dev/test interactions. 
Table~\ref{tab:data_stats} shows data statistics.
The system uses a SQL database of $27$ tables and $162$K entries. 
$96.6\%$ of  the queries require at least one join, and $93\%$ at least two joins. 
The most related work on ATIS to ours is \citet{Miller:96}, which we discuss in Section~\ref{sec:related}.

The most related corpora to ATIS are SCONE~\cite{Long:16context} and SequentialQA~\cite{Iyyer:17seq-qa}. 
SCONE~\cite{Long:16context} contains micro-domains consisting of stack- or list-like elements. The formal representation is linguistically-motivated and the majority of queries include a single binary predicate. All interactions include five turns. 
SequentialQA~\cite{Iyyer:17seq-qa} contains sequences of questions on a single Wikipedia table. 
Interactions are on average $2.9$ turns long, and were created by re-phrasing a question from a context-independent corpus~\cite{Pasupat:15compositional}.
In contrast, ATIS uses a significantly larger database, requires generating complex queries with multiple joins, includes longer interactions, and was collected through interaction with users. 
The supplementary material contains  analysis of the contextual phenomena observed in ATIS.

\paragraph{Pre-processing}

We pre-process the data to identify and anonymize entities (e.g., cities), numbers, times, and dates. 
We use string matching heuristics to identify entities and numbers, and identify and resolve times and dates using UWTime~\cite{Lee:14}. 
When resolving dates we use the original interaction date as the document time.
The supplementary material details this process.

\paragraph{Metrics}
We evaluate using query accuracy, strict denotation accuracy, and relaxed denotation accuracy. 
Query accuracy is the percentage of predicted queries that match the reference query. 
Strict denotation accuracy is the percentage of predicted queries that execute to  exactly the same table as the reference query. 
In contrast to strict, relaxed gives credit to a prediction query that fails to execute if the reference table is empty. 
In cases when the utterance is ambiguous and there are multiple gold queries,  we consider the query or table correct if they match any of the gold labels.

\paragraph{Systems}

We evaluate four systems: (a) \textsc{seq2seq-0}: the baseline encoder-decoder model (Section~\ref{sec:base_model}); (b) \textsc{seq2seq-h}: encoder-decoder with attention on current and previous utterances (Section~\ref{sec:multiple_utterance_attention}); (c) \textsc{s2s+anon}:  encoder-decoder with attention on previous utterances and anonymization scoring (Section~\ref{sec:anon}); and (d) \textsc{Full}: the complete approach including segment copying (Section~\ref{sec:snippets}). 
For \textsc{Full}, we evaluate with predicted and gold (\textsc{Full-Gold}) previous queries, and without attention on previous utterances (\textsc{Full-0}). 
All models except $\textsc{seq2seq-0}$ and \textsc{Full-0} use $\historylength=3$ previous utterances. 
We limit segment copying to segments that appear in the most recent query only.\footnote{While we only use segments from the most recent query, they often appear for the first time much earlier in the interaction, which influences their absolute position value $\snippetage$.} 
Unless specifically ablated, all experiments use pre-processing. 

\section{Results}
\label{sec:results}

\begin{table}[t]
\begin{footnotesize}
\begin{center}
\begin{tabular}{|l|c|c|c|} 
	\hline
	\multirow{ 2}{*}{\textbf{Model}} & \multirow{ 2}{*}{\textbf{Query}} & \multicolumn{2}{|c|}{\textbf{Denotation}} \\ \cline{3-4}
	& & \textbf{Relaxed} & \textbf{Strict} \\ 
	\hline
	\hline
	\multicolumn{4}{|l|}{\textbf{Development Results}} \\
	\hline
	\hline
    \textsc{seq2seq-0} & \stdev{28.7}{1.7} & \stdev{48.8}{1.4}  & \stdev{43.2}{1.8} \\ \dline
    \textsc{seq2seq-h} & \stdev{35.1}{2.2} & \stdev{59.4}{2.4} & \stdev{56.7}{3.2}  \\ \dline
    \textsc{s2s+anon} & \stdev{37.6}{0.7} & \stdev{61.6}{0.7} &  \stdev{60.6}{0.7} \\  \dline
    \textsc{Full-0} & \stdev{36.3}{0.5} & \stdev{61.5}{0.8} & \stdev{61.0}{0.9} \\ \dline
    \textsc{Full} & \stdev{37.5}{0.9} & \stdev{63.0}{0.7}&  \stdev{\mathbf{62.5}}{0.9} \\ \dline
   -- turn-level enc. & \stdev{37.4}{1.5} & \stdev{62.1}{2.5} & \stdev{61.4}{2.7} \\ \dline
   -- batch re-weight & \stdev{36.4}{0.6} & \stdev{61.8}{0.3} & \stdev{61.5}{0.4} \\ \dline
   -- input pos. embs. & \stdev{33.3}{0.2} & \stdev{57.9}{0.8} & \stdev{57.4}{0.8} \\ \dline
   -- query segments & \stdev{36.0}{0.9} & \stdev{59.5}{1.3} &  \stdev{58.3}{1.4} \\ \dline
   -- anon. scoring & \stdev{35.7}{0.5} & \stdev{60.8}{1.1} & \stdev{60.0}{1.0} \\ \dline
   -- pre-processing & \stdev{26.4}{6.1} & \stdev{53.3}{8.6} & \stdev{53.0}{8.5} \\ 
   \hline
   \hline
   \textsc{Full-Gold} & \stdev{42.1}{0.8} & \stdev{66.6}{0.7} & \stdev{66.1}{0.7} \\ \hline	
   \hline
	\multicolumn{4}{|l|}{\textbf{Test Results}} \\
	\hline
	\hline
	 \textsc{seq2seq-0} & \stdev{35.7}{1.5} & \stdev{56.4}{1.1} & \stdev{53.8}{1.0} \\ \dline
   \textsc{seq2seq-h} & \stdev{42.2}{2.0} & \stdev{66.6}{3.2} & \stdev{65.8}{3.4} \\ \dline
    \textsc{s2s+anon} & \stdev{44.0}{1.2} & \stdev{69.3}{1.0} & \stdev{68.6}{1.1} \\ \dline
    \textsc{Full-0} & \stdev{43.1}{1.3} & \stdev{67.8}{1.6} & \stdev{67.2}{1.6} \\ \dline
    \textsc{Full} & \stdev{43.6}{1.0} & \stdev{69.3}{0.8} &  \stdev{\mathbf{69.2}}{0.8} \\ \hline \hline
    \textsc{Full-Gold} & \stdev{47.4}{1.3} &  \stdev{72.3}{0.5} & \stdev{72.0}{0.5} \\ \hline	
\end{tabular}
\end{center}
\end{footnotesize}
\caption{Mean and standard deviation development and test results, including ablations on the \textsc{Full} model.}
\label{tab:results}
\end{table}

Table~\ref{tab:results} shows development and test results. 
We run each experiment five times and report mean and standard deviation. 
The main metric we focus on is strict denotation accuracy. 
The relatively low performance of \textsc{seq2seq-0}  demonstrates the need for context in this task. 
Attending on recent history significantly increases performance. 
Both \textsc{seq2seq} models score anonymized tokens as regular vocabulary tokens. 
Adding anonymized token scoring further increases performance (\textsc{s2s+anon}). 
\textsc{Full-0} and \textsc{Full} add segment copying and the turn-level encoder. 
The relatively high performance of \textsc{Full-0} shows that substituting segment copying with attention maintains and even improves the system effectiveness. 
However, the best performance is provided with \textsc{Full}, which combines both. 
This shows the benefit of redundancy in accessing contextual information. 
Unlike the other systems, both \textsc{Full} and \textsc{Full-0} suffer from cascading errors due to selecting query segments from previously incorrect predictions. 
The higher \textsc{Full-Gold} performance illustrates the influence of error propagation. 
While part of this error can be mitigated by having both attention and segment copying, this behavior is unlikely to be learned from supervised learning, where errors are never observed. 

Ablations show that all components contribute to the system performance. 
Performance drops when using a concatenation-based encoder instead of the turn-level encoder ($-$turn-level enc.; Section~\ref{sec:discourse_lstm}). 
Using batch-reweighting ($-$batch-reweight; Section~\ref{sec:learning}) and input position embeddings ($-$input pos. embs.; Section~\ref{sec:discourse_lstm}) are critical to the performance of the turn-level encoder. 
Removing copying of query segments from the interaction history lowers performance ($-$query segments; Section~\ref{sec:snippets}).
Treating indexed anonymized tokens as regular tokens, rather than using attention-based scoring and type embeddings, lowers performance ($-$anon. scoring; Section~\ref{sec:anon}).
Finally, pre-processing, which includes anonymization, is critical ($-$pre-processing). 

\definecolor{fullgoldcolor}{RGB}{230,25,75}
\definecolor{fullpredcolor}{RGB}{60,180,75}
\definecolor{seqseqbettercolor}{RGB}{0,130,200}
\definecolor{seqseqcolor}{RGB}{245,130,48}
\definecolor{seqseqonecolor}{RGB}{145,30,180}

\begin{figure}[t!]
\begin{center}
\begin{tikzpicture}
\footnotesize
\begin{groupplot}[group style={group size= 2 by 1},no markers,
		width=0.95\columnwidth,
        height=0.4\columnwidth,
       ymin=30, ymax=80,
        ytick={30,45,60,75},
        xlabel style={yshift=0ex,},
        legend style={at={(1.0,1.1)},anchor=south},
        legend columns=3,
        legend cell align={left},
        clip=false
]

\nextgroupplot[ylabel style={text width=0.3\columnwidth,align=center}, 
xlabel=(a) Interaction turns,
ylabel = Strict Denotation,
width=0.6\columnwidth, 
xmin=0,xmax=20,
        xtick={0,5,10,15,20},
        ylabel near ticks
]
       \addplot[draw=none,mark=none,name path=SeqSeqOneLower,forget plot] coordinates{
(0,64.62)
(1,40.98)
(2,40.49)
(3,39.14)
(4,35.07)
(5,33.03)
(6,37.35)
(7,37.28)
(8,39.59)
(9,35.23)
(10,35.30)
(11,34.67)
(12,33.69)
(13,41.25)
(14,35.20)
(15,34.03)
(16,31.76)
(17,31.48)
(18,44.28)
(19,31.54)
(20,34.32)
     };
     
     \addplot[draw=none,mark=none,name path=SeqSeqOneUpper,forget plot] coordinates{
(0,68.54)
(1,47.15)
(2,45.00)
(3,44.45)
(4,41.03)
(5,38.40)
(6,41.52)
(7,42.31)
(8,40.58)
(9,38.03)
(10,42.67)
(11,40.00)
(12,35.32)
(13,42.92)
(14,37.24)
(15,38.59)
(16,39.86)
(17,38.89)
(18,51.38)
(19,44.78)
(20,50.05)
 };     
     
     \addplot [fill=seqseqonecolor,opacity=0.2,forget plot] fill between[of=SeqSeqOneLower and SeqSeqOneUpper];
          \addplot[color=seqseqonecolor,style={ thick}] coordinates { 
       (0,66.58)
(1,44.06)
(2,42.74)
(3,41.79)
(4,38.05)
(5,35.71)
(6,39.43)
(7,39.80)
(8,40.09)
(9,36.63)
(10,38.99)
(11,37.33)
(12,34.51)
(13,42.08)
(14,36.22)
(15,36.31)
(16,35.81)
(17,35.19)
(18,47.83)
(19,38.16)
(20,42.19)
}; 
     \addlegendentry{\textsc{seq2seq-0}}
      
      \addplot[draw=none,mark=none,name path=FullPredLower,forget plot] coordinates{
(0,71.76)
(1,66.25)
(2,62.2)
(3,62.1)
(4,56.16)
(5,55.62)
(6,57.44)
(7,57.25)
(8,60.34)
(9,56.86)
(10,51.79)
(11,57.47)
(12,57.63)
(13,60.05)
(14,56.7)
(15,57.77)
(16,63.4)
(17,49.49)
(18,62.14)
(19,49.53)
(20,61.58)
     };
     
     \addplot[draw=none,mark=none,name path=FullPredUpper,forget plot] coordinates{
(0,73.82)
(1,68.05)
(2,66.13)
(3,65.52)
(4,62.24)
(5,59.99)
(6,62.04)
(7,58.22)
(8,67.24)
(9,62.74)
(10,61.55)
(11,60.39)
(12,66.31)
(13,67.28)
(14,64.12)
(15,70.81)
(16,69.57)
(17,55.69)
(18,68.29)
(19,68.36)
(20,68.42)
     };     
     
     \addplot [fill=fullpredcolor,opacity=0.2,forget plot] fill between[of=FullPredLower and FullPredUpper];
    
     \addplot[color=fullpredcolor, style={thick}] coordinates { 
(0,72.79)
(1,67.15)
(2,64.16)
(3,63.81)
(4,59.2)
(5,58.72)
(6,59.64)
(7,57.14)
(8,63.79)
(9,59.8)
(10,56.67)
(11,58.93)
(12,61.97)
(13,63.67)
(14,60.41)
(15,64.29)
(16,66.49)
(17,52.59)
(18,65.22)
(19,58.95)
(20,65.00)
}; 
     \addlegendentry{\textsc{Full}}

     \addplot[draw=none,mark=none,name path=SeqSeqLower,forget plot] coordinates{
(0,69.19)
(1,61.33)
(2,54.07)
(3,57.03)
(4,50.19)
(5,49.82)
(6,46.86)
(7,47.33)
(8,49.30)
(9,52.16)
(10,48.77)
(11,47.48)
(12,43.39)
(13,42.07)
(14,46.14)
(15,46.94)
(16,48.43)
(17,38.59)
(18,58.06)
(19,37.52)
(20,44.27)
     };
     
     \addplot[draw=none,mark=none,name path=SeqSeqUpper,forget plot] coordinates{
(0,74.39)
(1,67.83)
(2,64.66)
(3,62.68)
(4,55.29)
(5,51.66)
(6,52.90)
(7,54.17)
(8,58.63)
(9,58.33)
(10,57.90)
(11,54.38)
(12,58.02)
(13,55.93)
(14,59.98)
(15,66.40)
(16,60.76)
(17,50.30)
(18,61.94)
(19,57.21)
(20,63.23)
     };     
     
     \addplot [fill=seqseqcolor,opacity=0.2,forget plot] fill between[of=SeqSeqLower and SeqSeqUpper];
          \addplot[color=seqseqcolor,style={ thick}] coordinates { 
          (0,71.79)
(1,64.58)
(2,59.37)
(3,59.85)
(4,52.74)
(5,50.74)
(6,49.88)
(7,50.75)
(8,53.97)
(9,55.25)
(10,53.33)
(11,50.93)
(12,50.70)
(13,49.00)
(14,53.06)
(15,56.67)
(16,54.59)
(17,44.44)
(18,60.00)
(19,47.37)
(20,53.75)
}; 
     \addlegendentry{\textsc{seq2seq-h}}

           \addplot[draw=none,mark=none,name path=FullGoldLower,forget plot] coordinates{
(0,71.57)
(1,68.09)
(2,64.63)
(3,66.98)
(4,60.12)
(5,62.72)
(6,61.35)
(7,57.45)
(8,65.03)
(9,62.46)
(10,55.14)
(11,61.93)
(12,69.73)
(13,63.85)
(14,61.01)
(15,57.42)
(16,68.27)
(17,50.76)
(18,66.49)
(19,57.70)
(20,62.86)
     };
     
     \addplot[draw=none,mark=none,name path=FullGoldUpper,forget plot] coordinates{
(0,73.49)
(1,70.12)
(2,67.61)
(3,69.13)
(4,66.61)
(5,65.95)
(6,66.51)
(7,62)
(8,73.25)
(9,68.23)
(10,62.96)
(11,67.13)
(12,72.25)
(13,72.15)
(14,66.33)
(15,70.2)
(16,74.43)
(17,57.39)
(18,72.64)
(19,66.51)
(20,79.64)
     };     
     
     \addplot [fill=fullgoldcolor,opacity=0.2,forget plot] fill between[of=FullGoldLower and FullGoldUpper];
     
     \addplot[color=fullgoldcolor,style={ thick}] coordinates { 
(0,72.53)
(1,69.11)
(2,66.12)
(3,68.06)
(4,63.36)
(5,64.33)
(6,63.93)
(7,59.73)
(8,69.14)
(9,65.35)
(10,59.05)
(11,64.53)
(12,70.99)
(13,68)
(14,63.67)
(15,63.81)
(16,71.35)
(17,54.07)
(18,69.57)
(19,62.11)
(20,71.25)
      }; 
     \addlegendentry{\textsc{Full-Gold}}
     \addplot[draw=none,mark=none,name path=SeqSeqBetterLower,forget plot] coordinates{
(0,69.43)
(1,66.78)
(2,61.61)
(3,63.86)
(4,53.43)
(5,49.71)
(6,51.91)
(7,54.37)
(8,57.32)
(9,55.97)
(10,53.08)
(11,53.93)
(12,57.63)
(13,52.75)
(14,56.54)
(15,53.62)
(16,56.73)
(17,45.79)
(18,58.10)
(19,42.27)
(20,54.40)
     };
     
     \addplot[draw=none,mark=none,name path=SeqSeqBetterUpper,forget plot] coordinates{
(0,73.72)
(1,69.75)
(2,66.71)
(3,65.09)
(4,56.66)
(5,55.31)
(6,57.14)
(7,58.82)
(8,59.57)
(9,60.07)
(10,56.45)
(11,59.13)
(12,64.06)
(13,59.25)
(14,61.01)
(15,67.34)
(16,69.75)
(17,51.99)
(18,65.38)
(19,65.09)
(20,68.10)
     };     
     
     \addplot [fill=seqseqbettercolor,opacity=0.2,forget plot] fill between[of=SeqSeqBetterLower and SeqSeqBetterUpper];
     \addlegendentry{\textsc{s2s+anon}}
   
     \addplot[color=seqseqbettercolor,style={ thick}] coordinates { 
(0,71.58)
(1,68.27)
(2,64.16)
(3,64.48)
(4,55.04)
(5,52.51)
(6,54.52)
(7,56.60)
(8,58.45)
(9,58.02)
(10,54.76)
(11,56.53)
(12,60.85)
(13,56.00)
(14,58.78)
(15,60.48)
(16,63.24)
(17,48.89)
(18,61.74)
(19,53.68)
(20,61.25)
     }; 

\nextgroupplot[xlabel = (b) History length $\historylength$,
width=0.6\columnwidth,
xmin=-0.5,xmax=3.5,
xtick={0,1,2,3},
yticklabels={,,},
xshift=-0.7cm]

     \addplot[draw=none,mark=none,name path=SeqSeqLower,forget plot] coordinates{
     (0,41.4)
     (1,50.4)
     (2,54.5)
     (3,53.5)
     };
     
     \addplot[draw=none,mark=none,name path=SeqSeqUpper,forget plot] coordinates{
     (0,45)
     (1,54)
     (2,56.9)
     (3,59.9)
     };     
     
     \addplot [fill=seqseqcolor,opacity=0.2,forget plot] fill between[of=SeqSeqLower and SeqSeqUpper];

     \addplot[color=seqseqcolor,style={ thick}] coordinates { 
     (0,43.2) %
     (1,52.2) %
     (2,55.7) %
     (3,56.7) %
      }; 

     \addplot[draw=none,mark=none,name path=SeqSeqBetterLower,forget plot] coordinates{
     (0,49.1)
     (1,54.2)
     (2,59.4)
     (3,59.9)
     };
     
     \addplot[draw=none,mark=none,name path=SeqSeqBetterUpper,forget plot] coordinates{
     (0,50.5)
     (1,56.4)
     (2,60.0)
     (3,61.3)
     };     
     
     \addplot [fill=seqseqbettercolor,opacity=0.2,forget plot] fill between[of=SeqSeqBetterLower and SeqSeqBetterUpper];

     \addplot[color=seqseqbettercolor,style={ thick}] coordinates { 
     (0,49.8) %
     (1,55.3) %
     (2,59.7) %
     (3,60.6) %
     }; 

      \addplot[draw=none,mark=none,name path=FullPredLower,forget plot] coordinates{
        (0, 60.1)
        (1, 60.5)
        (2, 61.5)
        (3, 61.6)
     };
     
     \addplot[draw=none,mark=none,name path=FullPredUpper,forget plot] coordinates{
        (0, 61.9)
        (1, 61.9)
        (2, 63.3)
        (3, 63.4)
     };     
     
     \addplot [fill=fullpredcolor,opacity=0.2,forget plot] fill between[of=FullPredLower and FullPredUpper];

     \addplot[color=fullpredcolor, style={ thick}] coordinates { 
     (0,61.0) %
     (1,61.2) %
     (2,62.4) %
     (3,62.5) %
     }; 

     \addplot[draw=none,mark=none,name path=FullGoldLower,forget plot] coordinates{
        (0, 64.6)
        (1, 64.2)
        (2, 65.0)
        (3, 65.4)
     };
     
     \addplot[draw=none,mark=none,name path=FullGoldUpper,forget plot] coordinates{
        (0, 66.0)
        (1, 65.4)
        (2, 66.4)
        (3, 66.8)
     };     

     \addplot [fill=fullgoldcolor,opacity=0.2,forget plot] fill between[of=FullGoldLower and FullGoldUpper];
     
     \addplot[color=fullgoldcolor,style={ thick}] coordinates { 
     (0,65.3) %
     (1,64.8) %
     (2,65.7) %
     (3,66.1) %
     }; 
 \end{groupplot}

\end{tikzpicture}

\end{center}
\vspace{-8pt}
\caption{Mean development strict denotation accuracy as function of turns and $\historylength$.}
\label{fig:interaction_turn_accuracy}
\vspace{+5pt}
\end{figure}  

Figure~\ref{fig:interaction_turn_accuracy}(a) shows the performance as interactions progress. 
All systems show a drop in performance after the first utterance, which is always context-independent. 
As expected, \textsc{seq2seq-0} shows the biggest drop. 
The \textsc{Full} approach is the most stable as the interaction progresses. 

Figure~\ref{fig:interaction_turn_accuracy}(b) shows the performance as we decrease the number of previous utterances used for attention $\historylength$. 
Without the turn-level encoder and segment copying (\textsc{seq2seq-h} and \textsc{s2s+anon}), performance decreases significantly as $\historylength$ decreases. 
In contrast, the \textsc{Full} model shows a smaller decrease ($1.5\%$). 
The supplementary material includes attention analysis demonstrating the importance of previous-utterance attention.
However, attending on fewer utterances improves inference speed: \textsc{Full-0} is $30\%$ faster than \textsc{Full}.

\begin{table}[t]
\begin{footnotesize}
\begin{center}
\begin{tabular}{|l|c|c|c|} 
	\hline
	\multirow{ 2}{*}{\textbf{Model}} & \multirow{ 2}{*}{\textbf{Query}} & \multicolumn{2}{|c|}{\textbf{Denotation}} \\ \cline{3-4}
	& & \textbf{Relaxed} & \textbf{Strict} \\ 
	\hline
	\hline
	\multicolumn{4}{|l|}{\textbf{Development Results}} \\
	\hline
	\hline
	  \textsc{s2s+anon} & \stdev{44.4}{1.2} & \stdev{69.9}{0.3} & \stdev{68.9}{0.3} \\ \dline
    \textsc{Full} &  \stdev{42.8}{0.2} & \stdev{68.8}{0.2} &\stdev{68.4}{0.2} \\ \hline \hline
    \textsc{Full-Gold} & \stdev{47.5}{0.2} &  \stdev{71.5}{0.4} & \stdev{70.7}{0.6} \\ \hline
       \hline
	\multicolumn{4}{|l|}{\textbf{Test Results}} \\
	\hline
	\hline
	\textsc{s2s+anon} &\stdev{43.9}{0.3} & \stdev{67.4}{1.0} & \stdev{67.2}{1.0} \\ \dline
    \textsc{Full} & \stdev{44.3}{0.2} & \stdev{66.3}{0.3} & \stdev{66.3}{0.4} \\ \hline \hline
    \textsc{Full-Gold} & \stdev{47.2}{0.3} & \stdev{68.2}{0.5} & \stdev{67.9}{0.4} \\ \hline
\end{tabular}
\end{center}
\end{footnotesize}
\caption{Results on the original split of the data.}
\label{tab:original_split}
\end{table}

Finally, while we re-split the data due to scenario sharing between train and test early in development and used this split only for development, we also evaluate  on the original split (Table~\ref{tab:original_split}). 
We report mean and standard deviation over three trials. 
The high performance of \textsc{s2s+anon} potentially indicates it benefits more from the differences between the splitting procedures.

\section{Analysis}
\label{sec:analysis}

We analyze errors made by the full model on thirty development interactions.
When analyzing the output of \textsc{Full}, we focus on error propagation and analyze predictions that resulted in an incorrect table when using \textsc{Full}, but a correct table when using \textsc{Full-Gold}. 
$56.7\%$ are due to selection of a segment that contained an incorrect constraint.
$43.4\%$ of the errors are caused by a necessary segment missing during generation.
$93.0\%$ of all predictions are valid SQL and follow the database schema.
We also analyze the errors of \textsc{Full-Gold}. We observe that $30.0\%$ of errors are due to generating constraints that were not mentioned by the user.
Other common errors include generating relevant constraints with incorrect values ($23.3\%$) and missing constraints ($23.3\%$).

We also evaluate our model's ability to recover long-distance references while constraints are added, changed, or removed, and when target attributes change.
The supplementary material includes the analysis details. 
In general, the model resolves references well. However, it fails to recover constraints mentioned in the past following a user focus state change~\cite{Grosz:86}.

\section{Discussion}

We study models that recover context-dependent executable representations from user utterances by reasoning about interaction history.
We observe that our segment-copying models  suffer from error propagation when extracting segments from previously-generated queries.
This could be mitigated by training a model to ignore erroneous segments, and recover by relying on attention for generation. 
However, because supervised learning does not expose the model to erroneous states, a different learning approach is required. 
Our analysis demonstrates that our model is relatively insensitive to interaction length, and is able to recover both explicit and implicit references to previously-mentioned entities and constraints. 
Further study of user focus change is required, an important phenomenon that is relatively rare in ATIS.

\section*{Acknowledgements}

This research was supported by the NSF (CRII-1656998), Schmidt Sciences, a gift from Google, and cloud computing credits from Amazon. 
We thank Valts Blukis, Luke Zettlemoyer, and the anonymous reviewers for their helpful comments.

\balance
\bibliography{main}
\bibliographystyle{acl_natbib}

\clearpage
\nobalance
\appendix

\section{Data Processing}
\label{sec:appendix:preprocessing}
\paragraph{Date and Number Resolution} 
We replace instances of spelled-out multi-digit numbers in the original data (e.g., \nlstring{flight two three five}) with a numerical representation (e.g., \nlstring{flight 235}).
We resolve expressions containing references to dates using UWTime~\cite{Lee:14}.
Interactions in ATIS are annotated with the date they took place, which we use as document time. 
We use the \textit{newswire} annotator in UWTime to annotate each utterance in ATIS with date tags, and add one week to any predicted dates that occur before the document date. 
This heuristic follows the assumption that users always ask for information in the future.
UWTime is able to predict dates contained in the gold queries in the training data with approximately $70\%$ accuracy.
Without using interaction dates and resolving date expressions, the model would not be able to generate correct dates, unless through overfitting to the training set.
Previous work on ATIS addresses the problem of resolving relative date expressions by modifying the output queries.
For example, \citet{Zettlemoyer:09} add logical constants such as \textit{tomorrow} to the lambda-calculus lexicon.
To the best of our knowledge, no previous context-dependent work on ATIS uses interaction dates to recover the referents of date expressions.

\paragraph{Entity Anonymization}
We replace known entities in user utterances and SQL queries with anonymized tokens. 
Using the database, we generate a set of known entities and their natural language and SQL forms, for example the set of city names from the \sqlstring{city} table.
When anonymizing an example, we first identify entities that occur in the input sequence, and replace each with a unique anonymized token, using the same token for entities that occur multiple times.
We also identify numerical constants, which include flight numbers and times, as entities in the input sequence.
This process gives us a set of entities in the input sequence and their corresponding anonymized tokens.
During training, we replace any tokens in the gold SQL query that are in this set with the appropriate anonymized token.
Entity anonymization is separate of entity scoring, described in Section~\ref{sec:anon}, which computes scores for generating anonymized tokens independently of generating raw SQL tokens.
When entity scoring is ablated in our experiments, entity anonymization is still applied as a pre-processing technique, but anonymized tokens with indices are used as regular members of the input and output vocabularies.

\paragraph{Post-processing}
After generating a query but before evaluating it against the SQL database or comparing with the gold labels, we post-process it.
We de-anonymize all generated anonymized tokens using the anonymization set extracted from the input sequences, and correct mismatched parentheses by adding closing parenthesis at the end of the query. 

\section{Copying Segments}\label{sec:appendix:segments}
\paragraph{Segment Extraction from Previous Queries}
To construct $\snippetset(\query)$, we deterministically extract subtrees of the SQL parse tree from $\query$, as well as \sqlstring{SELECT} statements containing special modifiers like $\sqlstring{MIN}$.
We use the \texttt{sqlparse} package to construct a tree, and recursively traverse its subtrees. 
We consider all subtrees of a \sqlstring{WHERE} clause.
We separate children of conjunctions into distinct subtrees.
When evaluating with predicted queries, we extract segments from the most recent prediction that has correct syntax and follows the database structure. 

\paragraph{Alignment of Segments with Gold Queries}
During training, we align the set of extracted segments $\snippetset(\query)$ with the gold query to construct a new query that contains references to extracted segments.
We first extract known entities, e.g. city names, from the current utterance $\utterancename_\utteranceindex$, using the entity set described above.
We greedily substitute the longest extracted segments in $\snippetset(\query)$ first.
If segment $\snippetname$ is a subsequence of the gold query $\query_\utteranceindex$, we replace that subsequence with a reference to it.
We do not replace the subsequence if it contains one of the entities extracted from the input sequence; entities mentioned in the current utterance should be explicitly generated in the current prediction, as these are most likely not references to previous constraints.

\section{Implementation Details}\label{sec:appendix:implementation}
\paragraph{Learning}
We use the \textsc{Adam} optimizer~\cite{Kingma:14adam} with an initial global learning rate of $0.001$. 
The batch size $B = 16$.
We use patience as a stopping mechanism, with an initial patience of $10$ epochs.
We compute loss and token-level accuracy on a held-out validation set after each epoch.
This set includes $5\%$ of the training data, and when using our split of the dataset, does not contain scenarios that are present in the remaining $95\%$ of the data.
We use the same validation set across all of our experiments.
After an epoch, if token-level loss on the validation set increases since the previous epoch, the global learning rate is multiplied by  $0.8$.
When token-level accuracy on the validation set increases to a maximum value during training, we multiply the current patience by $1.01$.
During training, we apply dropout with probability $0.5$ after the first decoder LSTM layer, and after computing $\mathbf{m}_k$ at each decoding step.
The model parameters used to evaluate on the development and test sets are those that yielded the highest string-level validation accuracy.

During training, when multiple gold labels are present, we use the shortest.
Loss is not computed for utterance-query pairs if, after pre-processing and query segment alignment, the gold label contains more than $200$ tokens.
However, if using the turn-level encoder, this pair's input sequence is encoded and the turn-level state is updated.
During evaluation on the development and test sets, we limit generation to $300$ tokens.

If not using the turn-level encoder, we delimit the previous and current utterances with a special delimiter token when encoding the inputs (Section~\ref{sec:multiple_utterance_attention}).
The corresponding encoder hidden states of the delimiters are not used during attention.

\paragraph{Parameters}
We use word embeddings of size $400$.
The word embeddings are not pre-trained. 
Utterance age embeddings are of size $50$.
Query segment age embeddings are of size $64$.
For all models that use query segment copying, $\maxsnippetage=4$.
All LSTMs have a hidden size of $800$.
The sizes of the learned matrices are: $\attentionbilinear \in \mathbb{R}^{850 \times 800}$, $\intermediateweights \in \mathbb{R}^{1650 \times 800}$, $\outputvocabweights \in \mathbb{R}^{800 \times \length{\outputvocab}}$, $\outputvocabbias_w \in \mathbb{R}^{\length{\outputvocab}}$, and $\snippetweights \in \mathbb{R}^{800 \times 1600}$.
Unless otherwise noted, the initial hidden state and cell memory of all LSTMs are zero-vectors.
All parameters are initialized randomly from a uniform distribution $U[-0.1,0.1]$.

\section{Results}

\subsection{Overfitting in Original Data Split}\label{sec:original_split_overfitting}
\begin{table}[t]
\begin{footnotesize}
\begin{center}
\begin{tabular}{|l|l|l|} 
	\hline
	\textbf{Split} & \textbf{Model} & \textbf{Strict Denotation} \\ \hline
	\multirow{ 2}{*}{Original} & \textsc{Full}& \stdev{68.4}{0.2} \\ \ddline{2-3}
	& -- preprocess. & \stdev{67.1}{1.0} \textbf{(-1.3)} \\ \hline
	\multirow{ 2}{*}{Ours} & \textsc{Full}& \stdev{62.5}{0.9} \\ \ddline{2-3}
	& -- preprocess. & \stdev{53.0}{8.5} \textbf{(-9.5)} \\ \hline
	\end{tabular}
\end{center}
\end{footnotesize}
\caption{Strict table accuracy results using our model with and without pre-processing on the development sets of the original data split and our data split.\vspace{+1.5em}}
\label{tab:split_dev_results}
\end{table}
We assess overfitting on the training set of the original split of the data by measuring how performance changes when data pre-processing is removed.
Table~\ref{tab:split_dev_results} shows that removing data pre-processing lowers the performance of \textsc{Full} by around $9\%$ when using our data split.
However, on the original split of the data, performance drops by only $1.3\%$. This relatively high performance is only possible due to learned biases within the scenarios. 

\subsection{Ablations using Gold Previous Queries}
\begin{table}[t]
\begin{footnotesize}
\begin{center}
\begin{tabular}{|l|c|c|c|} 
	\hline
	\multirow{ 2}{*}{\textbf{Model}} & \multirow{ 2}{*}{\textbf{Query}} & \multicolumn{2}{|c|}{\textbf{Denotation}} \\ \cline{3-4}
	& & \textbf{Relaxed} & \textbf{Strict} \\ \hline
   \textsc{Full-Gold} & \stdev{42.1}{0.8} & \stdev{66.6}{0.7} & \stdev{66.1}{0.7} \\ \dline
   -- turn-level encoder &\stdev{42.5}{1.7} & \stdev{66.3}{1.7} & \stdev{65.7}{1.9} \\ \dline
   -- batch re-weighting &\stdev{41.1}{0.7} &\stdev{65.5}{0.3} & \stdev{64.8}{0.6} \\ \dline
   -- input pos. embs. & \stdev{38.0}{0.4} &  \stdev{61.4}{1.1} & \stdev{60.5}{1.1} \\ \dline
   -- anon. scoring & \stdev{40.8}{0.7} & \stdev{64.7}{1.4} & \stdev{63.9}{1.3}\\ \dline
   -- pre-processing & \stdev{35.3}{6.9} &\stdev{57.9}{8.4} & \stdev{57.4}{8.2} \\ \hline
	\end{tabular}
\end{center}
\end{footnotesize}
\caption{Ablations on \textsc{Full-Gold}, showing performance on the development set averaged over all utterances. Gold queries are provided for previous query segment extraction. In each model, $\historylength=3$. We show average and standard deviation over five trials for each model.\vspace{+0.5em}}
\label{tab:ablation_results_gold}
\end{table}
Table~\ref{tab:ablation_results_gold} shows results on ablating components from \textsc{Full} when extracting segments from previous gold queries.

\section{Analysis}
\subsection{Data Analysis}\label{sec:appendix:data_analysis}

\definecolor{constraintellipsiscolor}{RGB}{230,25,75}
\definecolor{targetellipsiscolor}{RGB}{60,180,75}
\definecolor{coreferencecolor}{RGB}{169,77,234}

\begin{figure}[t!]
\begin{center}
\begin{tikzpicture}
 \begin{axis}[
        width=1\columnwidth,
        height=0.55\columnwidth,
        font=\footnotesize,
        xmin=0,xmax=10,
        ymin=0, ymax=60,
        xtick={0,2,4,6,8,10},
        ytick={0,10,20,30,40,50,60},
        bar width=12pt,
        xlabel style={yshift=0ex,},
        xlabel=Age of reference,
        ylabel=\% of non-first utterances,
        legend pos=north east,
        legend cell align={left}]

\addplot[color=constraintellipsiscolor,style=thick] coordinates { 
(1,58.1)
(2,38.5)
(3,29.7)
(4,26.4)
(5,14.2)
(6,6.8)
(7,2.7)
(8,4.7)
(9,2)
(10,2)
}; 
\addlegendentry{Ellipsis on constraints}
 \addplot[color=targetellipsiscolor,style=thick] coordinates { 
(1,4.1)
(2,3.4)
(3,1.4)
(4,0.7)
(5,0)
(6,0)
(7,0)
(8,0)
(9,0)
(10,0)
}; 
\addlegendentry{Ellipsis on target}
\addplot[color=coreferencecolor,style=thick] coordinates { 
(1,7.4)
(2,2.7)
(3,0.7)
(4,0)
(5,0)
(6,0)
(7,0)
(8,0)
(9,0)
}; 
\addlegendentry{Referring expression}
\end{axis}
\end{tikzpicture}
\end{center}
\caption{Distribution of referent ages over sampled utterances in the development set.}
\label{fig:reference_age_analysis}
\end{figure}  

The two most common  contextual phenomena in ATIS are ellipsis and referring expressions.
Ellipsis refers to utterances that omit relevant information mentioned in the past.
For example, $\query_2$ in Figure~\ref{fig:convsql} contains flight endpoint constraints that are not present in $\utterancename_2$.
To recover these constraints, the model must resolve an implicit reference to $\utterancename_1$.
Referring expressions explicitly refer to earlier constraints or system responses.
In the example utterances, the phrase \nlstring{which ones} in $\utterance_3$ refers to the table returned by executing $\query_2$.

An analysis of  twenty development-set interactions shows that utterances after the first turn contain on average $1.85$ omitted constraints, and $60.1\%$ of utterances after the first turn omit at least one constraint.
Figure~\ref{fig:reference_age_analysis} shows the distribution of referent ages over all utterances in the twenty analyzed interactions from the development set, considering three major types of references: ellipsis on constraints, ellipsis on the attributes targeted by the user request (e.g., if a user omits the target attribute in the current utterance, but it is clear from context), and referring expressions (explicit mentions of previous constraints). 
$9\%$ of utterances after the first utterance omit the target attribute; $11\%$ contain a referring expression to previous results or constraints.

\subsection{Attention Analysis}
\label{sec:appendix:attention}

Figures~\ref{fig:attention_full}, \ref{fig:attention_short}, and~\ref{fig:attention_s2s} demonstrate the robustness of the full approach when attending over multiple previous utterances.
Each figure shows attention while processing the same example, which is the third utterance in an interaction from the development set.
This interaction exemplifies a user focus state change.
To process the third utterance, \nlstring{show flights}, correctly, the model must be able to recover constraints mentioned in the first utterance, although the user briefly changes focus in the second utterance.
In each figure, the query is shown on the lefthand side, and the utterances are at the bottom. 
The opacity of each cell represents the attention probability for each token during each generation step.
Darker lines in the figure separate the three utterances.

Figure~\ref{fig:attention_full} shows the attention computed by \textsc{Full} when provided with gold query segments. 
The decoder attends over entities in the previous utterance, including the flight endpoints and date, when generating the query.
Figure~\ref{fig:attention_short} shows the attention computed by \textsc{Full-0} when provided with gold query segments.
This model does not have explicit access to the constraints mentioned in the first utterance. 
It is unable to recover these constraints, and instead makes up constraints, such as endpoints and a date, while also making a semantic mistake, \texttt{city.city\_name = 1200}.
This demonstrates the robustness provided by the attention mechanism in the full model.
For comparison, Figure~\ref{fig:attention_s2s} shows the attention computed by \textsc{s2s+anon} on the same example. 
Like \textsc{Full}, this model attends over entities in the first utterance, including flight endpoints and the date. 
Both \textsc{Full} and \textsc{s2s+anon} were able to recover the correct query.

Figures~\ref{fig:attention_full_2}, \ref{fig:attention_short_2}, and~\ref{fig:attention_s2s_2} demonstrate how the ability to copy query segments is critical to our model's performance. 
\textsc{Full} and \textsc{Full-0} are able to recover the correct query. 
In both cases, \sqlstring{SEGMENT\_9}, which is extracted from the previous query, contains the flight endpoint constraints (from the first utterance), as well as a constraint that the flight be the shortest one available (from the second utterance).
\textsc{s2s+anon} is unable to recover the minimum-time constraint, even though it has the ability to attend over the relevant tokens in the second utterance.
The ability of \textsc{Full-0} to recover this constraint without attending on previous utterances demonstrates the benefit that copying previous segments provides.
These examples also show that when copying query segments, fewer decoding steps are required.

\subsection{Contextual Analysis}
\label{sec:appendix:context_analysis}

We construct several example interactions targeting the contextual phenomena discussed in Section~\ref{sec:analysis}, and test \textsc{Full} against them.
\textbox{
$\utterancename_1$: show me flights from seattle to denver after 6am\\
$\utterancename_2$: leaving after 7am \\
$\utterancename_3$: leaving after 8am \\
$\utterancename_4$: leaving before 9am \\
$\utterancename_5$: leaving after 10am \\
$\utterancename_6$: leaving from san francisco
} %
This example shows ellipsis of both constraints (flight endpoints) and target attribute (flights) while modifying existing constraints. 
\textsc{Full} is able to predict correct queries for all new utterances as the user continues to change constraints and elided values increase in age.
Whether $\query_6$ includes a time constraint or not is ambiguous; \textsc{Full} generates a query to search for all flights from San Francisco to Denver regardless of time.
\textbox{$\utterancename_1$: show me flights from seattle to denver \\
$\utterancename_2$: leaving after 7am \\
$\utterancename_3$: stopping in san francisco \\
$\utterancename_4$: on american airlines \\
$\utterancename_5$: which is the cheapest \\
$\utterancename_6$: with breakfast} 
This example shows ellipsis of both constraints and target attribute while adding constraints. 
\textsc{Full} is able to predict correct queries for all utterances in this interaction.
\textbox{$\utterancename_1$: show me flights from seattle to denver after 6am\\
$\utterancename_2$: how much does it cost \\
$\utterancename_3$: what meal is offered \\
$\utterancename_4$: which airlines are available \\
$\utterancename_5$: what type of airplane does it use \\
$\utterancename_6$: what ground transportation is available}
This example shows ellipsis on constraints (endpoints and time) while changing the target attribute. 
Additionally, it demonstrates change in focus while the user switches from asking for flights to asking about airlines.
While ambiguous, the final utterance $\utterancename_6$ is interpreted as finding ground transportation in Denver.
\textsc{Full} is able to recover correct queries for all utterances in this interaction.
\textbox{$\utterancename_1$: show me flights from seattle to denver after 6am \\
$\utterancename_2$: what ground transportation is available in denver \\
$\utterancename_3$: i want to fly on delta}
This example shows ellipsis when user focus changes, temporarily rendering the origin city and time constraints irrelevant.
\textsc{Full} predicts queries for the first two utterances correctly, but fails to generate the correct origin city constraint in the third prediction.
Instead, it generates a constraint that Denver is the origin city.
When $\utterance_2$ is removed from the interaction, both predictions are correct.

\begin{figure*}
\centering
\include{attention_full_1}
\caption{Attention computed by \textsc{Full} after a user state focus change. Compare to Figures~\ref{fig:attention_short} and \ref{fig:attention_s2s}.}
\label{fig:attention_full}
\end{figure*}

\begin{figure*}
\centering
\include{attention_short_1}
\caption{Attention computed by \textsc{Full-0} when generating a  query for an utterance after a user state focus change. The model does not have explicit access to flight endpoint or date constraints, and is unable to generate the correct query. Figure~\ref{fig:attention_full} shows \textsc{Full} on this example; Figure~\ref{fig:attention_s2s} shows \textsc{s2s-anon}.}
\label{fig:attention_short}
\end{figure*}

\begin{figure*}
\centering
\include{attention_s2s_1}
\caption{Attention computed by \textsc{s2s+anon} after a user state focus change. Compare to Figures~\ref{fig:attention_full} and \ref{fig:attention_short}.}
\label{fig:attention_s2s}
\end{figure*}

\begin{figure*}
\centering
\include{attention_full_2}
\caption{Attention computed by \textsc{Full}, demonstrating copying of query segments. Figure~\ref{fig:attention_short_2} shows \textsc{Full-0} on this example; Figure~\ref{fig:attention_s2s_2} shows \textsc{s2s-anon}.}
\label{fig:attention_full_2}
\end{figure*}

\begin{figure*}
\centering
\include{attention_short_2}
\caption{Attention computed by \textsc{Full-0} that makes use of query segments to recover constraints it does not otherwise have explicit access to. Figure~\ref{fig:attention_full_2} shows \textsc{Full} on this example; Figure~\ref{fig:attention_s2s_2} shows \textsc{s2s-anon}.}
\label{fig:attention_short_2}
\end{figure*}

\begin{figure*}
\centering
\include{attention_s2s_2}
\caption{Attention computed by \textsc{s2s-anon}, demonstrating a failure in recovering constraints. Figure~\ref{fig:attention_full_2} shows \textsc{Full} on this example; Figure~\ref{fig:attention_short_2} shows \textsc{Full-0}.}
\label{fig:attention_s2s_2}
\end{figure*}

\end{document}